%% file: arxiv.tex
\documentclass{article} % For LaTeX2e
\usepackage{preprint,times}

\input{math_commands.tex}

\usepackage{hyperref}
\usepackage{url}
\usepackage{booktabs}
\usepackage{amssymb}
\usepackage{xspace}
\usepackage{threeparttable}
\usepackage{graphicx}
\usepackage{wrapfig}
\usepackage{pifont}

\makeatletter
\DeclareRobustCommand\onedot{\futurelet\@let@token\@onedot}
\def\@onedot{\ifx\@let@token.\else.\null\fi\xspace}

\def\eg{\emph{e.g}\onedot} 
\def\ie{\emph{i.e}\onedot} 
 
\def\etc{\emph{etc}\onedot}

\makeatother

\title{UniLat3D: Geometry-Appearance Unified \\Latents for Single-Stage 3D Generation}

\author{
    \textbf{Guanjun Wu}\textsuperscript{1,2*} \hspace{0.1cm}
    \textbf{Jiemin Fang}\textsuperscript{1*$\dagger$} \hspace{0.1cm}
    \textbf{Chen Yang}\textsuperscript{1*} \hspace{0.1cm}
    \textbf{Sikuang Li}\textsuperscript{1,3} \hspace{0.1cm}
    \textbf{Taoran Yi}\textsuperscript{1,2} \\
    \textbf{Jia Lu\textsuperscript{1,2}} \hspace{0.1cm}
    \textbf{Zanwei Zhou\textsuperscript{1,3}} \hspace{0.1cm} 
    \textbf{Jiazhong Cen\textsuperscript{1,3}} \hspace{0.1cm}
    \textbf{Lingxi Xie\textsuperscript{1}} \hspace{0.1cm}
    \textbf{Xiaopeng Zhang\textsuperscript{1}} \\
    \textbf{Wei Wei\textsuperscript{2}} \hspace{0.1cm}
    \textbf{Wenyu Liu\textsuperscript{2}} \hspace{0.1cm}
    \textbf{Xinggang Wang\textsuperscript{2}} \hspace{0.1cm} 
    \textbf{Qi Tian\textsuperscript{1$\dagger$}} \\
    \textsuperscript{*}Equal Contribution. \quad
    \textsuperscript{$\dagger$}Corresponding Authors.\\
    \textsuperscript{1}Huawei Inc. 
    \textsuperscript{2}Huazhong University of Science and Technology 
    \textsuperscript{3}Shanghai Jiaotong University \\[0.2cm]
}

\definecolor{frenchblue}{rgb}{0.0, 0.45, 0.73}
\newcommand{\comments}[1]{\textcolor{frenchblue}{#1}}

\finalcopy % Uncomment for camera-ready version, but NOT for submission.

\begin{document}

\maketitle

\begin{figure}[h]
\vspace{-15pt}
\includegraphics[width=\linewidth]{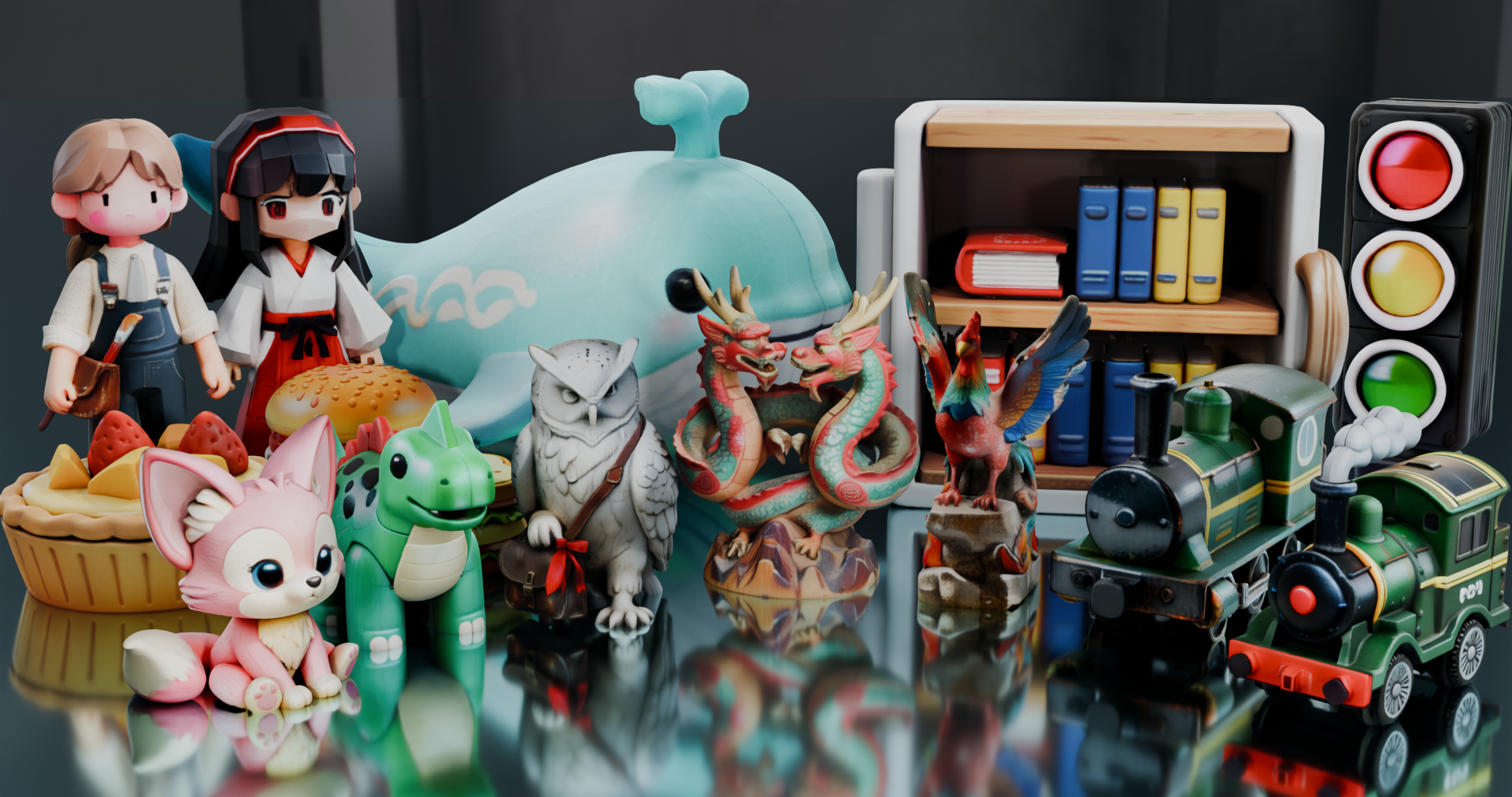}
\vspace{-15pt}
\caption{Gallery of UniLat3D. Our method generates high quality 3D assets in seconds. }
\end{figure}

\begin{abstract}
High-fidelity 3D asset generation is crucial for various industries. While recent 3D pretrained models show strong capability in producing realistic content, most are built upon diffusion models and follow a two-stage pipeline that first generates geometry and then synthesizes appearance. Such a decoupled design tends to produce geometry–texture misalignment and non-negligible cost. In this paper, we propose \textbf{UniLat3D}, a unified framework that encodes geometry and appearance in a single latent space, enabling direct single-stage generation. Our key contribution is a geometry–appearance Unified VAE, which compresses high-resolution sparse features into a compact latent representation -- \textbf{UniLat}. UniLat integrates structural and visual information into a dense low-resolution latent, which can be efficiently decoded into diverse 3D formats, \eg, 3D Gaussians and meshes. Based on this unified representation, we train a single flow-matching model to map Gaussian noise directly into UniLat, eliminating redundant stages. Trained solely on public datasets, UniLat3D produces high-quality 3D assets in seconds from a single image, achieving superior appearance fidelity and geometric quality. More demos and code are available at \url{https://unilat3d.github.io/}

\end{abstract}

\section{Introduction}
3D content generation has witnessed rapid growth in recent years, becoming an increasingly essential capability across various applications, including game/film production, virtual/augmented reality, industrial design, and embodied AI. Recent advances in 3D generative frameworks~\citep{zhang2024clay,hunyuan3d2025hunyuan3d2.1,lai2025hunyuan3d2.5,yang2024hunyuan3d1.0,zhao2025hunyuan3d2.0,trellis,li2025step1x,hong2023lrm,gslrm2024,ma20254dlrm,ren2024l4gm,zou2024triplane} have demonstrated impressive progress in synthesizing vivid and realistic 3D assets, while some approaches~\citep{li2024craftsman,wu2024direct3d,wu2025direct3ds2,chen2025ultra3d,li2025sparc3d,ye2025hi3dgen} dive into accurate geometry and fine-grained shape generation.

Despite this rapid progress, the majority of recent high-quality 3D generation frameworks are diffusion-based, and typically adopt a multi-stage design: geometry is generated first, followed by texture or appearance synthesis.
This paradigm, rooted in the conventional separation of geometry and appearance, has been adopted by both latent-based pipelines~\citep{trellis} and mesh-based frameworks~\citep{li2025step1x,hunyuan3d2025hunyuan3d2.1}, remaining the prevailing design but entailing inherent drawbacks. First, the separate generation introduces an inevitable gap between geometry and appearance, potentially leading to misalignment with the target 3D asset. Second, the two-stage process introduces additional computation budget, \textit{e.g.}, current mesh-based methods~\citep{hunyuan3d2025hunyuan3d2.1} first generate the geometry, and then synthesize the corresponding texture based on both the condition image and geometry generated in the first stage.
Notably, the research trajectory in both vision and graphics~\citep{mildenhall2020nerf,kerbl3Dgaussians} has long favored unification over separation -- just as object detection evolved from multi-stage Faster R-CNN~\citep{girshick2015fastrcnn} to single-stage YOLO~\citep{redmon2016yolo}. We aim to create a similar unification of geometry and appearance generation, which is expected to offer more convenience and possibilities for exploring 3D generation under a more extensible and unified framework.

\begin{wrapfigure}{r}{0.6\textwidth}
    \centering
    \vspace{-10pt}
    \includegraphics[width=\linewidth]{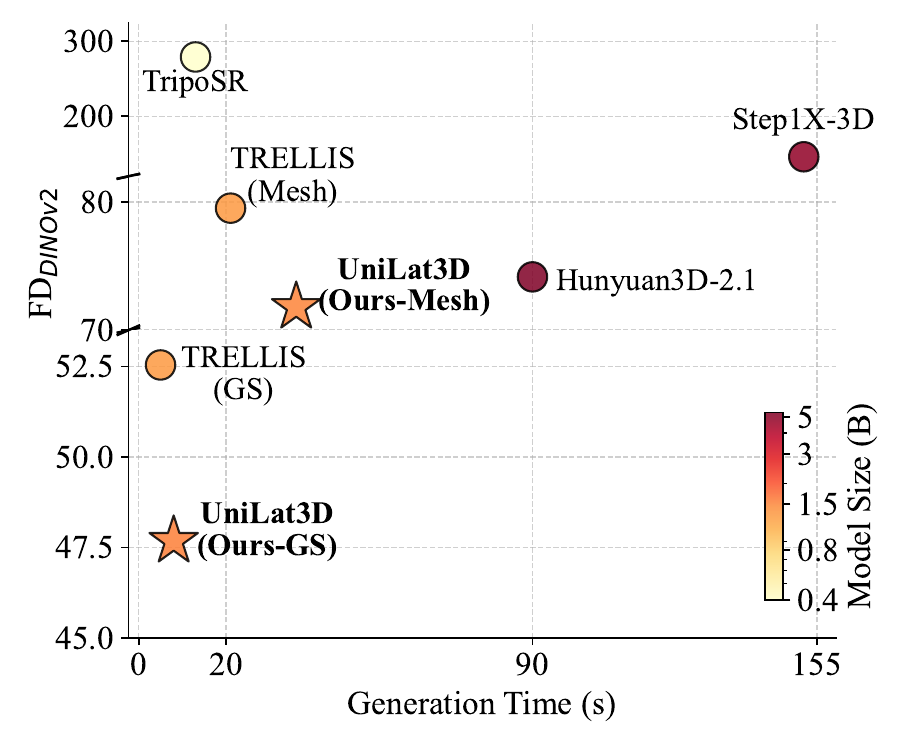}
    \vspace{-20pt}
    \caption{Evaluation on Toys4K~\citep{toys4k}. Colors stand for model sizes. Lower generation time and smaller $\text{FD}_\text{DINOv2}$ indicate better performance, \ie the left bottom corner.}
    \label{fig:right}
\end{wrapfigure}
To this end, we introduce a unified 3D representation that inherently encodes geometry and appearance in a single latent space, enabling direct single-stage generation. Our key insight is that such a representation is naturally aligned—free from geometry–texture mismatches—and highly efficient, as it avoids redundant intermediate steps. Inspired by TRELLIS~\citep{trellis}, we first transform the 3D asset into sparse structured features. A unified variational autoencoder, UniVAE, is designed to compress high-resolution sparse features into a compact latent space, termed \textbf{UniLat}. The UniLat can then be efficiently upsampled and sparsified back onto high-resolution latents that serve as a universal basis for decoding into various renderable 3D representations, such as 3D Gaussians~\citep{kerbl3Dgaussians} and meshes. Thanks to the simplicity and expressive design of UniLat, we are able to, for the first time, achieve single-stage 3D generation through one flow-matching model that maps cubic Gaussian noise directly into the geometry-appearance unified latents. Beyond efficiency, UniLat also offers strong extensibility, which can serve as a versatile 3D prior that can be seamlessly integrated into large multimodal models, facilitating cross-modal understanding and generation.  
Our method, \textbf{UniLat3D}, trained only on publicly available datasets, achieves superior appearance fidelity while maintaining strong geometric accuracy, demonstrating the effectiveness of unifying geometry and appearance within a single-stage paradigm.

Our contributions are summarized as follows.
\begin{itemize}
\setlength{\itemsep}{0pt}
      \item We propose a novel framework, UniLat3D, which bridges the gap between geometry and appearance by a single diffusion model in high-quality 3D generation.
      \item A novel UniLat representation is introduced by encoding geometry and appearance into a unified latent space, ensuring high-efficiency feature fusion.
      \item As in Fig.~\ref{fig:right}, extensive experiments demonstrate UniLat3D's state-of-the-art performance. We expect our framework to pave a novel way for exploring 3D generation in a more unified and scalable paradigm.
\end{itemize}

\section{Related Works}
\subsection{3D Generation by Lifting 2D Diffusion Models}
Lifting 2D diffusion models to 3D has been an effective but challenging approach. DreamFusion~\citep{poole2022dreamfusion} proposes Score Distillation Sampling (SDS) to distill knowledge from the 2D diffusion model into a radiance field. \cite{tang2023dreamgaussian,yi2023gaussiandreamer,yi2024gaussiandreamerpro,yin20234dgen,ren2023dreamgaussian4d,liu2024sherpa3d,wang2023prolificdreamer} follow this methodology to generate high-quality 3D Gaussians~\citep{kerbl3Dgaussians} in minutes. Meanwhile, \cite{jain2022zero,liu2023zero1to3,shi2023mvdream,huang2024mv,long2023wonder3d,liu2023one,yang2024gaussian} fine-tune the image diffusion model to generate multi-view consistent images for synthesizing 3D assets. Video diffusion models~\citep{yang2024cogvideox,yu2024viewcrafter,xing2024dynamicrafter,ren2025gen3c,zhao2024genxd,gao2024cat3d,wu2025cat4d,liang2024diffusion4d} are also explored to synthesize high-quality 3D/4D representations~\citep{wu20244d,yang2023real,zhang2024dynamic2dgs,zhang2025togs}. However, most of these methods need iterative optimization from different views in each generation process, which takes a non-negligible cost, while hallucination may appear, \eg, Janus phenomenon, due to the lack of 3D priors.

\subsection{3D Generation by Pretraining 3D Foundation Models}
With the emergence of large-scale 3D datasets, \eg, Objaverse~\citep{deitke2023objaverse}, 3D foundation models have been constructed and pretrained to have strong reconstruction and generation abilities.

\paragraph{3D Foundation Reconstruction Models.} Some feed-forward 3D reconstruction methods~\citep{wang2024dust3r,wang2025vggt,zhang2024monst3r,smart2024splatt3r,li2025megasam,wang2025continuous,yang2025fast3r}, using vision Transformer~\citep{visiontransformer} (VIT) to encode and match input images' features and recover their relative 3D poses, depths, semantics~\citep{sun2025uni3r,xu2025uniugg}, and other 3D information~\citep{jiang2025gausstr,smart2024splatt3r}. Those methods achieve nearly real-time reconstruction given an image sequence, while maintaining accurate pose/depth estimation, and high-quality novel view synthesis.

\paragraph{3D Foundation Generation Models.} A series of 3D foundation models aims to generate high-quality 3D representations with few or a single image(s) as input in seconds. In the early stage, 3D Generation mainly focuses on structure\&shape generation~\citep{ren2024xcube,vahdat2022lion} or other latent representation~\citep{yang2024atlas}. Point-E~\citep{nichol2022point} trains a 3D diffusion model, which is used for generating point clouds from text/image prompts.
VecSet~\citep{zhang20233dshape2vecset} proposes to encode 3D assets into vector representations, which are further applied in the geometry diffusion models~\citep{chen2025ultra3d,hunyuan3d2025hunyuan3d2.1,lai2025hunyuan3d2.5,zhang2024clay,li2024craftsman,xiong2025octfusion}. Then, texture diffusion models~\citep{hunyuan3d2025hunyuan3d2.1,li2025step1x} are followed to color the high-quality mesh.
TRELLIS~\citep{trellis} and some recent works~\citep{ye2025hi3dgen,wu2025direct3ds2,li2025sparc3d,chen2025ultra3d} encode multiview images into sparse 3D voxel representations and then decode them into high-quality 3D assets.
Several methods are proposed to generate dynamic objects~\citep{chen2025v2m4,zhang2025gaussian,wu2025animateanymesh} or extend 3D generation to the part level~\citep{chen2025partgen,dong2025one,chen2025autopartgen,yang2025omnipart}.

We observe that most 3D diffusion models split the generation process into two phases -- geometry and appearance. Our research aims to bridge the gap between geometry and appearance in 3D generation by introducing a unified latent space while maintaining the strong performance of 3D diffusion models.

\section{Preliminary}
Recently, TRELLIS~\citep{trellis}, a powerful 3D generation framework, has enabled generating high-quality 3D assets in seconds. This is achieved by proposing sparse structured latents (SLATs) $\mathbf{z}_\mathrm{slat}$ to represent the 3D asset, which can be decoded into different 3D representations. 

\paragraph{Sparse Structured Latent Representation.} 
SLAT is defined as a series of latents located at activated surface voxels of the 3D asset, which can be formulated as $\mathbf{z}_\mathrm{slat} = \{z_i, p_i\}_{i=1}^L$, where $z_i \in R^c$ is a $c$-dimensional latent at the voxel position $p_i \in R^3$, $i=\{1,2,...L\}$, $N$ denotes the grid resolution and $L << N^3$. 
The coordinates $\{p_i\}$, representing coarse geometry, are computed by voxelizing the 3D asset. The latents $\{z_i\}$, representing appearance and detailed geometry\footnote{Some detailed geometry properties will be decoded from latents $\{z_i\}$, \eg 3D Gaussian positions and mesh vertices. This will be denoted as `appearance' for short in the following content.}, are obtained by aggregating and encoding visual features $\mathbf{f}=\{f_i, p_i\}_{i=1}^L$, extracted by a vision encoder~\citep{oquab2023dinov2} from multiple views of the asset. 
To learn geometry and appearance respectively, TRELLIS constructs two separate VAE models, \ie, geometry VAE $\{\mathcal{E}_\mathrm{geo},\mathcal{D}_\mathrm{geo}\}$ and appearance VAE $\{\mathcal{E}_\mathrm{app}, \mathcal{D}_\mathrm{app}\}$.

Specifically, the encoder of the geometry VAE transforms activated voxels $\mathbf{p} = \{p_i\}$ to geometry latents $\mathbf{z}_\mathrm{geo} \in R^{\frac{N}{s} \times \frac{N}{s} \times \frac{N}{s} \times c}$ with a downsampling factor $s$:
\begin{equation}
    \mathbf{z}_\mathrm{geo} = \mathcal{E}_\mathrm{geo}(\mathbf{p} ); \;\; \mathbf{p} = \mathcal{D}_\mathrm{geo}(\mathbf{z}_\mathrm{geo}).
\end{equation}
The sparse appearance VAEs encodes the sparse 3D features $\mathbf{f}$ into SLATs $\mathbf{z}_\mathrm{slat}$, and decodes SLATs into 3D representations $\mathcal{O}$ as:
\begin{equation}\label{eq:trellis_slat_encoding}
    \mathbf{z}_\mathrm{slat} = \mathcal{E}_\mathrm{app}(\mathbf{f}); \;\; \mathcal{O} = \mathcal{D}_\mathrm{app}(\mathbf{z}_\mathrm{slat}).
\end{equation}
Note that $\mathcal{E}_\mathrm{app}$ only converts $\mathbf{f}$ in the feature dimension. The coordinate information is modeled by $\mathcal{E}_\mathrm{geo}$ individually.

\paragraph{Sparse Structured Latent Generation.} To generate SLAT $\mathbf{z}_\mathrm{slat}$, TRELLIS proposes a two-stage generation pipeline. Given the condition image $\mathbf{I}$, TRELLIS builds a geometry generation flow Transformer $\mathcal{F}_\mathrm{geo}$ to synthesize geometry latents $\mathbf{z}_\mathrm{geo}$ from the noise $\mathbf{\epsilon}$. Then, the activated voxels $\mathbf{p}$ can be decoded by $\mathcal{D}_\mathrm{geo}$:
\begin{equation}
    \mathcal{F}_\mathrm{geo}:(\mathbf{\epsilon},t,\mathbf{I}) \rightarrow \mathbf{z}_\mathrm{geo}; \;\; \mathbf{p} = \mathcal{D}_\mathrm{geo}(\mathbf{z}_\mathrm{geo}),
\end{equation}
where $t$ is the denoising timestep. After that, the appearance noise can be added to the activated voxels $\mathbf{p}$ to get the structured noise $\epsilon_\mathrm{app} = \{ \epsilon_i, p_i\}$. The sparse appearance flow Transformer is optimized to predict $\mathbf{z}_\mathrm{slat}$, and the final 3D representation $\mathcal{O}$ can be computed by the appearance decoder $\mathcal{D}_\mathrm{app}$:
\begin{equation}
    \begin{gathered}
    \mathcal{F}_\mathrm{app}:(\mathbf{\epsilon}_\mathrm{app},t,\mathbf{I}) \rightarrow \mathbf{z}_\mathrm{slat}; \;\; \mathcal{O} = \mathcal{D}_\mathrm{app}(\mathbf{z}_\mathrm{slat}).
    \end{gathered}
\end{equation}

\section{Method}

\subsection{Overall Framework}

\paragraph{Geometry-Appearance Unified Latent Representation.} 
Different from TRELLIS~\citep{trellis}, which obtains sparse structured latents $\mathbf{z}_\mathrm{slat} = \{z_i, p_i\}_{i=1}^L$ in two separate stages, we propose a dense compressed \textbf{Lat}ent representation with geometry and appearance \textbf{Uni}fied (UniLat) $\mathbf{z}_\mathrm{uni} \in R^{M \times M \times M \times d}$ which can be obtained in one single stage, where $d$ is the number of unified latent's channels, $M=\frac{N}{V}$, and $V$ denotes the compression ratio. In the reconstruction stage, we construct a UniLat variational autoencoder (Uni-VAE) $\{\mathcal{E}_\mathrm{uni}, \mathcal{D}_\mathrm{uni,\{gs,mesh\}}\}$ to encode the 3D assets efficiently. The rich geometry and appearance of an assets $\mathcal{O}$ can be encoded into the UniLat $\mathbf{z}_\mathrm{uni}$, which can be further decoded into 3D representations via decoder $\mathcal{D}_\mathrm{uni}$ as:
\begin{equation}
\mathbf{z}_\mathrm{uni} \leftarrow \mathcal{E}_\mathrm{uni}(\mathcal{O}); \;\; \mathcal{O}=\mathcal{D}_\mathrm{uni}(\mathbf{z}_\mathrm{uni}).
\end{equation}

The unified decoder $\mathcal{D}_\mathrm{uni}$ is composed of a upsampling block $\mathcal{D}_\mathrm{up}$ and 3D representation decoders $\mathcal{D}_\mathrm{gs,mesh}$. For more details, please refer to Sec.~\ref{subset:decoder}.

\paragraph{Geometry–Appearance Unified Latent Generation.}
With geometry and appearance already fused in our UniLat representation $\mathbf{z}_\mathrm{uni}$, the generation process becomes naturally streamlined. A unified generative model $\mathcal{F}_\mathrm{uni}$ is employed to directly denoise compact noises $\epsilon$ into UniLat $\mathbf{z}_\mathrm{uni}$, which can then be decoded by $\mathcal{D}_\mathrm{uni}$ into the desired 3D representation:
\begin{equation}\label{eq:flow_unify}
    \mathcal{F}_\mathrm{uni}:(\mathbf{\epsilon},t,\mathbf{I}) \rightarrow \mathbf{z}_\mathrm{uni};  \;\;  \mathcal{O} = \mathcal{D}_\mathrm{uni}(\mathbf{z}_\mathrm{uni}).
\end{equation}

\subsection{UniLat Variational Autoencoder}
\begin{figure}[t]
\begin{center}
%\framebox[4.0in]{$\;$}
\includegraphics[width=\linewidth]{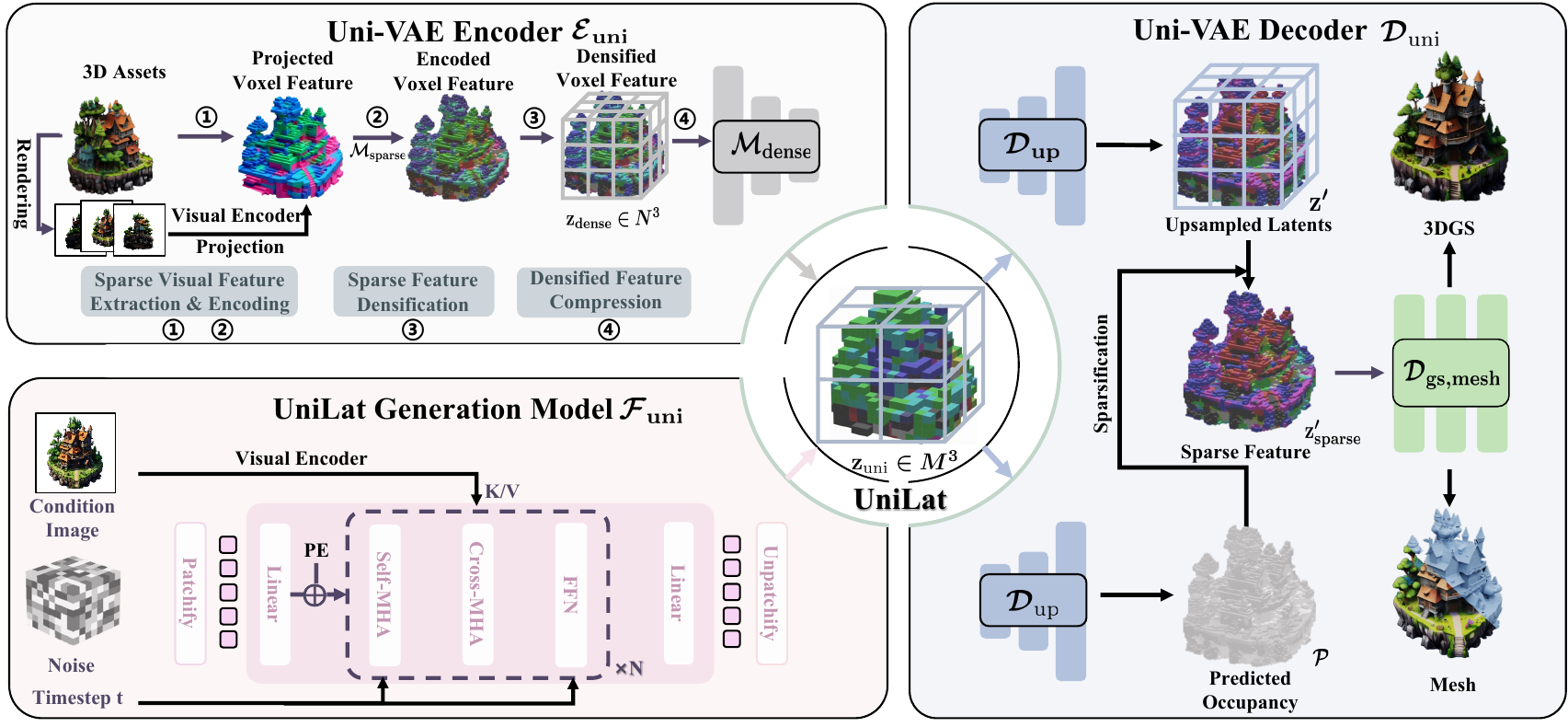}
\end{center}
\caption{Illustration of the UniLat3D framework. In the reconstruction stage, the encoder of Uni-VAE $\mathcal{E}_\mathrm{uni}$ converts the 3D asset $\mathcal{O}$ to the unified latent -- UniLat $\mathbf{z}_\mathrm{uni}$, which can be directly denoised from noise $\epsilon$ by a single flow model $\mathcal{F}_\mathrm{uni}$ in the generation stage. The obtained UniLat can be transformed into target 3D representations by the decoder $\mathcal{D}_\mathrm{uni}$.}
\end{figure}

\subsubsection{Encoder}
We design an encoder $\mathcal{E}_\mathrm{uni}$ to convert various 3D assets into our UniLats. $\mathcal{E}_\mathrm{uni}$ consists of several key stages: sparse visual feature extraction, sparse appearance encoding, sparse visual feature densification, and densified feature compression. These stages are supported by two core modules: the sparse appearance encoding module $\mathcal{M}_\mathrm{sparse}$ and the dense feature compression module $\mathcal{M}_\mathrm{dense}$. 
% The encoding process begins with the high-dimensional voxelized 3D features $\mathbf{f} = {f_i,p_i}$, derived from a 3D asset $\mathcal{O}$, and proceeds through the aforementioned stages to obtain the unified latent $\mathbf{z}_\mathrm{uni}$.
% The UniLat $\mathbf{z}_\mathrm{uni}$ obaining begins with the high dimensional voxelized 3D features $\mathbf{f} = \{f_i,p_i\}$. 
The encoding process begins with converting a 3D asset $\mathcal{O}$ to sparse visual features $\mathbf{f} = \{f_i,p_i\}$, following the multi-view visual feature projection proposed in TRELLIS~\citep{trellis}.
Then we employ sparse appearance feature module $\mathcal{M}_\mathrm{sparse}$ to get $\mathbf{z}_\mathrm{sparse}$ by $\mathbf{z}_\mathrm{sparse} = \mathcal{M}_\mathrm{sparse}(\mathbf{f)}$. These two stages are named sparse visual feature extraction and sparse appearance encoding, respectively. Later, we introduce the sparse feature densification process to fill the empty space in the sparse latents and get $\mathbf{z}_\mathrm{dense}$.
As computation on $\mathbf{z}_\mathrm{dense}$ is expensive,  we perform the densified feature compression phase, which encodes the processed features into lower-resolution compact latents, \ie UniLat $\mathbf{z}_\mathrm{uni}$. Finally, the UniLat decoder $\mathcal{D}_\mathrm{uni}$ upsamples the compressed unified latents $\mathbf{z}_\mathrm{uni}$ back onto high-resolution 3D representations, supporting both 3D Gaussian and mesh outputs.

\comments{
}

\paragraph{Sparse Feature Densification.}
For the sparse feature $\mathbf{z}_\mathrm{sparse}$ with appearance encoded, the geometry is given by indicating which location is empty.
To merge both geometry and appearance information into unified latents $\mathbf{z}_\mathrm{uni}$,  the structured appearance latents $\mathbf{z}_\mathrm{sparse} = \{(z_\mathrm{sparse,i}, p_i)\}_{i=1}^{L}$ are converted to dense features $\mathbf{z}_\mathrm{dense}$.  All the empty space is assigned with zero features $\{0,p_j\}_{j \neq i}^{N^3-L}$. Then, the sparse structured latents can be transformed to dense unified latents:
\begin{equation}
    \mathbf{z}_\mathrm{dense} : \{\mathbf{z}_\mathrm{dense}[p_i] = \mathbf{z}_\mathrm{sparse,i};\mathbf{z}_\mathrm{dense}[p_j] = 0\}.
\end{equation}
Here, $\mathbf{z}_\mathrm{dense}$ is a set of compact dense latents that includes the whole space information.

\paragraph{Densified Feature Compression.}
We use $\mathcal{M}_\mathrm{dense}$ to encode both the geometry and appearance features. Similar to 2D/2.5D diffusion models~\citep{dcae,blattmann2023stable}, $\mathcal{M}_\mathrm{dense}$ downsamples $\mathbf{z}_\mathrm{dense} \in N^3$ to \textbf{UniLats} $\mathbf{z}_\mathrm{uni} \in M^3$ with a downsampling factor $s$:

\begin{equation}
    \mathbf{z}_\mathrm{uni} = \mathcal{M}_\mathrm{dense}(\mathbf{z}_\mathrm{dense}).
\end{equation}

The geometry and appearance features are further fused by the downsampling encoding process, ensuring rich information in the UniLat $\mathbf{z}_\mathrm{uni}$ at the low resolution.         

\subsubsection{Decoder}
\label{subset:decoder}
Uni-Decoder $\mathcal{D}_\mathrm{uni}$ includes two modules: upsampling block $\mathcal{D}_\mathrm{up}$ and 3D representation decoders $\mathcal{D}_\mathrm{
{gs,mesh}}$. The high-resolution dense coordinate and features $\mathbf{z}_\mathrm{dense}^\prime \in R^{N^3\times (C+1)}$ are computed by $\mathcal{D}_\mathrm{up}$, then the pruning process is performed on the dense features $\mathbf{z}_\mathrm{dense}^\prime$ to obtain sparse features $\mathbf{z}_\mathrm{sparse}^\prime $. Finally, representation decoders $\mathcal{D}_\mathrm{gs,mesh}$ output the final 3D representations.

\paragraph{Latent Upsampling and Sparsification.}
Given a compact but low-resolution UniLat $\mathbf{z}_\mathrm{uni}$, the core challenge is to reconstruct high-quality 3D assets in a detailed manner. To address this, we introduce an upsampling block that lifts $\mathbf{z}_\mathrm{uni}$ to higher-resolution latents. Leveraging our geometry–appearance unified representation, we can simultaneously predict voxel occupancy, which guides a pruning step to remove redundant regions among the upsampled latents. This yields a sparse set of high-resolution latents that retain both efficiency and fidelity. 

Given UniLat $\mathbf{z}_\mathrm{uni} \in R^{M^3\times d}$, our proposed upsampling blocks $\mathcal{D}_\mathrm{up}$ compute the appearance and geometry features at resolution $N$ as :
\begin{equation}
    \mathbf{z}_\mathrm{dense}^\prime = \mathcal{D}_\mathrm{up}(\mathbf{z}_\mathrm{uni}).
\end{equation}
Note that both $\mathbf{z}_\mathrm{dense}^\prime = \{\mathcal{P} \in R^{N^3\times 1},\mathbf{z}^{\prime } \in N^3\times c \}$ are high-resolution dense features. Note that directly performing computation on $\mathbf{z}_\mathrm{dense}$ is expensive, so we propose to prune the low-importance area to enhance efficiency. The sparse features $\mathbf{z}_\mathrm{sparse}$ are filtered with a signed function:
\begin{equation}
    \mathbf{z}^\prime_\mathrm{sparse} : \{ \mathbf{z}^\prime_i,p_i \mid \mathcal{P}[p_i] > 0  \},
\end{equation}

\paragraph{3D Representation Decoders.}
Two 3D representation decoders are designed to transform the pruned latents into renderable 3D outputs, \textit{i.e.}, 3D Gaussians and meshes. Both decoders share a backbone of sparse Transformer blocks, similar to TRELLIS, but differ in their task-specific output heads. For 3D Gaussians, the decoder $\mathcal{D}_\mathrm{gs}$ maps the latent $\mathbf{z}_\mathrm{uni}$ to attributes of 3D Gaussian primitives $\mathcal{O}_\mathrm{gs}$ using sparse Transformer blocks and 3D linear projection layers. An additional occupancy head is employed to predicts voxel occupancy, enabling direct supervision of the reconstructed geometry. 

\begin{figure}[t]
  \centering
  \includegraphics[width=0.9\linewidth]{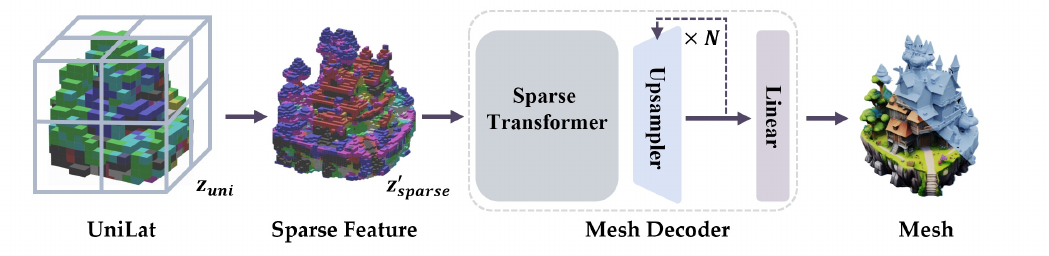}
    \vspace{-1em}
  \caption{Mesh decoder architecture.}
  \label{fig:decoder_mesh}
\end{figure}

For meshes, similar to Gaussian decoder, we first upsample $\mathbf{z}_\mathrm{uni}$ and perform sparsification; the resulting $\mathbf{z}^\prime_{\mathrm{sparse}}$ is processed by a stack of sparse Transformer blocks. As illustrated in Fig.~\ref{fig:decoder_mesh}, hierarchical upsampling is then applied to progressively increase the feature resolution: each stage performs octree-style subdivision, where each voxel is divided into eight sub-voxels to double the spatial resolution along each axis, followed by residual sparse 3D convolutions that refine local features and preserve gradient flow during training. In practice, three such blocks increase the resolution from $64^3$ to $512^3$. After upsampling, a sparse linear output layer predicts SDF values, voxel-corner deformations, and interpolation weights (\textit{i.e.}, the SparseFlex~\citep{he2025sparseflex} parameters), from which we extract mesh vertices and faces efficiently. To enable multi-scale geometry supervision and reduce computational overhead, occupancy prediction heads are attached at each resolution and supervised with corresponding voxel-level occupancy.

To scale $\mathcal{D}_\mathrm{mesh}$ to higher resolutions, we adopt a pruning strategy that removes voxels entirely outside or inside object boundaries, thereby reducing computational overhead. We further introduce a detail augmentation strategy, where depth and normal maps are rendered from zoomed-in camera views with a differentiable rasterizer, enabling the decoder to learn fine-grained surface details from localized partial observations. With these techniques, UniLat3D produces meshes at a resolution of $512^3$, doubling the resolution achieved by TRELLIS.

\subsection{UniLat Generation Model}
With Uni-VAE, we construct a generation model $\mathcal{F}_\mathrm{uni}$ based on rectified flow matching to denoise compact noise $\epsilon$ into condition-followed UniLats $\mathbf{z}_\mathrm{uni}$. A single flow Transformer model $\mathcal{F}_\mathrm{uni}$ with full attention layers is built to predict the velocity at timestamp $t$ under the noise level as $v =\mathcal{F}_\mathrm{uni}(\mathbf{x}_\mathrm{uni}, t, I)$ and $\mathbf{x}_\mathrm{uni}$ denotes the denoised noise $\epsilon$ and timestamp $t$. The whole flow Transformer optimization process follows the diffusion guidance given condition $\mathbf{I}$ with its condition encoder.
The latent features with both geometry and appearance information are denoised. The obtained UniLat $\mathbf{z}_\mathrm{uni}$ can be directly fed into the representation decoder $\mathcal{D}_\mathrm{uni}$ to predict the final 3D representation $\mathcal{O}$.

\subsection{Optimization}

\paragraph{Uni-VAE.}  We use both geometry and appearance supervision to train the $\mathcal{D}_\mathrm{uni}$.
Following TRELLIS~\citep{trellis}, we joint optimize $\mathcal{E}_\mathrm{uni}$ and $\mathcal{D}_\mathrm{gs}$ with the following loss: 
\begin{equation}
    \mathcal{L} = \lambda_\mathrm{l1}\mathcal{L}_\mathrm{l1} + \lambda_\mathrm{lpips}\mathcal{L}_\mathrm{lpips} + \lambda_\mathrm{ssim}\mathcal{L}_\mathrm{ssim} + \lambda_\mathrm{kl}\mathcal{L}_\mathrm{kl} + \lambda_\mathrm{dice}\mathcal{L}_\mathrm{dice} +
    \lambda_\mathrm{reg}\mathcal{L}_\mathrm{reg}.
\end{equation}
$\mathcal{L}_\mathrm{l1}$ denotes the L1 color loss, and $\mathcal{L}_\mathrm{lpips}$ and $\mathcal{L}_\mathrm{ssim}$ stand for inception-based losses. $\mathcal{L}_\mathrm{kl}$ is employed for optimizing $\mathcal{E}_\mathrm{uni}$. $\mathcal{L}_\mathrm{dice}$ and $\mathcal{L}_\mathrm{reg}$ are used to supervise geometry and decoded representations. 

For the mesh decoder, we adopt a hierarchical supervision aligned with the multi-scale upsampling described in Sec.~\ref{subset:decoder}. Occupancy prediction heads are attached at each resolution, and are trained with corresponding voxel-level occupancy targets.
% \[
% \mathcal{L}_{\mathrm{occ}}=\sum_{s\in\{128,256,512\}}\mathcal{L}_{\mathrm{dice}}^{(s)}.
% \]
The overall mesh objective is
\begin{equation}
\label{eq:loss_mesh}
\mathcal{L}_{\mathrm{mesh}}=\lambda_{\mathrm{geo}}\mathcal{L}_{\mathrm{geo}}+\lambda_{\mathrm{color}}\mathcal{L}_{\mathrm{color}}+\lambda_{\mathrm{reg}}\mathcal{L}_{\mathrm{reg}}+\lambda_{\mathrm{occ}}\mathcal{L}_{\mathrm{occ}},
\end{equation}
where $\mathcal{L}_{\mathrm{geo}}$, $\mathcal{L}_{\mathrm{color}}$, and $\mathcal{L}_{\mathrm{reg}}$ follow TRELLIS.
To alleviate the computational cost at high resolutions, training proceeds in two stages. \textit{Stage-1} optimizes $\mathcal{D}_\mathrm{mesh}$ up to $256^3$ resolution using Eq.~\eqref{eq:loss_mesh}. \textit{Stage-2} introduces an independent $256\!\to\!512$ upsampling block with its own pruning head; this new block is optimized while the Stage-1 pathway remains frozen. After decoding, lightweight post-processing removes invisible or degenerate faces and fills small holes.

\paragraph{Rectified Flow Models.} Once the Uni-VAE has been trained, all the UniLat $\mathbf{z}_\mathrm{uni}$ are predicted by the Uni-Encoder $\mathcal{E}_\mathrm{uni}$. For optimizing the rectified flow Transformer, we mainly follow the CFM Loss. Given encoded latents $\mathbf{x}_\mathrm{uni}$ and noise $\mathbf{\epsilon}$, we minimize the objective function $\mathcal{L}_\mathrm{CFM}$~\citep{lipman2022flowmatching} as:
\begin{equation}
L_\mathrm{CFM}(\theta) = \mathbb{E}_{t, x_0, \epsilon} \| v(\mathbf{x}_\mathrm{uni}, t) - (\mathbf{\epsilon} - \mathbf{z}_\mathrm{uni}) \|_2^2.
\end{equation}

\section{Experiments}
\subsection{Implementation Details}
Our framework is implemented in PyTorch~\citep{paszke2019pytorch} and built upon the open-source project TRELLIS~\citep{trellis}. FlashAttention-3\citep{shah2024flashattention} is employed to accelerate Transformer training, yielding a $1.5\times$ speedup. 
Both VAE and flow models are trained on 64 GPUs within two weeks. 

\paragraph{Uni-VAE.} To accelerate and stabilize Uni-VAE training, we initialize $\mathcal{E}_\mathrm{sparse}$ and $\mathcal{D}_\mathrm{sparse}$ with the pretrained weights from TRELLIS. During the first 240k iterations, only $\mathcal{E}_\mathrm{dense}$ and $\mathcal{D}_\mathrm{up}$ are optimized, after which the entire Uni-VAE is trained end-to-end for an additional 90k iterations following TRELLIS. For the mesh decoder, we freeze $\mathcal{D}_\mathrm{uni}$ and train our high-resolution mesh decoder from scratch. Unless otherwise specified, Adam~\citep{kingma2014adam} is used with a learning rate of $1\times 10^{-4}$.

\paragraph{UniLat Flow Transformer.} For training the rectified flow models, we adopt DINOv3~\citep{simeoni2025dinov3} as the image encoder and apply classifier-free guidance~\citep{ho2022classifierCFG} with a drop rate of 0.1. The model is first trained for 500k iterations with a batch size of 256 and a learning rate of $1\times 10^{-4}$, and then fine-tuned for 160k iterations with a batch size of 1024 and a learning rate of $1\times 10^{-5}$.

\subsection{Experimental Setup}

\paragraph{Training Datasets.} 
UniLat3D is trained exclusively on publicly available datasets. Following the data preparation pipeline of TRELLIS~\citep{trellis}, we curate and process approximately 450k high-quality 3D assets from Objaverse (XL)~\citep{deitke2023objaverse}, ABO~\citep{collins2022abo}, 3D-FUTURE~\citep{3dfuture}, and HSSD~\citep{khanna2023hssd}. To enable occupancy supervision at multiple scales, we perform voxelization at each resolution. Additional details on data preprocessing can be found in~\citep{trellis}.

\paragraph{Evaluation Datasets.} The evaluation is performed on two datasets. One is the whole Toys4K~\citep{toys4k} dataset, including 3218 high-quality 3D assets, which is also used in the previous method~\citep{trellis}. However, we observe that many samples of Toys4K tend to have simple geometry or appearance details. We construct a more complex dataset for comprehensive evaluation, including 500 high-quality assets collected from the Sketchfab platform and 500 assets sampled from Toys4K.
Condition images for qualitative comparisons and user studies are collected from~\cite{chen2025ultra3d,wu2025direct3ds2} or generated via VLMs.

\paragraph{Evaluation Setups.} For VAE reconstruction evaluation, we use the PSNR, SSIM, and LPIPS metrics. For appearance generation quality, we compute the CLIP~\citep{radford2021learning} score -- similarity between rendered images and condition images, and FD (Fréchet distance)~\citep{heusel2017gansFDscore} measured by DINOv2~\citep{oquab2023dinov2} on 4 views of each generated asset and ground truth images. 
We evaluate and compare our method with recent SOTA 3D generation models, \ie Hunyuan3D-2.1~\citep{hunyuan3d2025hunyuan3d2.1}, TRELLIS~\citep{trellis}, Step1X-3D~\citep{li2025step1x}, TripoSR~\citep{tochilkin2024triposr} for image-conditioned generation, and Stable3DGen~\citep{ye2025hi3dgen} and Direct3D-S2~\citep{wu2025direct3ds2} for geometry generation quality comparison. We report Uni3D~\citep{zhou2023uni3d} and ULIP~\citep{xue2023ulip} metrics for mesh geometry quality. 
% The Blender rendering pipeline is adopted for all generated mesh assets. 
% \paragraph{Blender Rendering Setups.} 
For mesh rendering, we mainly use Blender~\citep{blender} as a mesh renderer to render high-quality images. We set FOV=40, render resolution=512, and set normalization to each loaded object.

\begin{figure}[t]
\centering
\includegraphics[width=\linewidth]{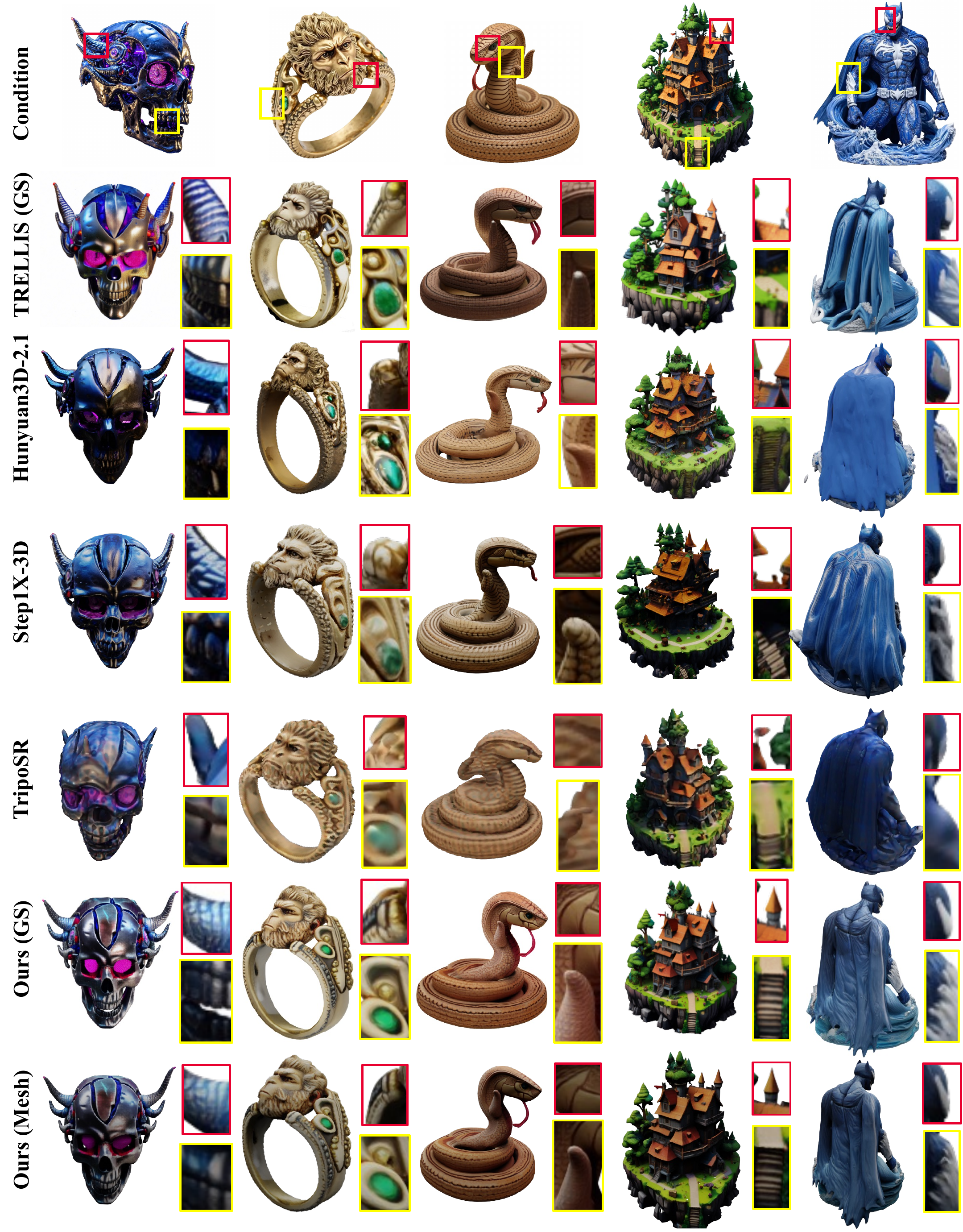}
\caption{Qualitative comparisons with other methods. Thanks to our unified representation, UniLat3D achieves superior performance and better correspondence with input images.}
\label{fig:comparison_opensource}
\end{figure}

\begin{figure}[t]
\centering
\includegraphics[width=\linewidth]{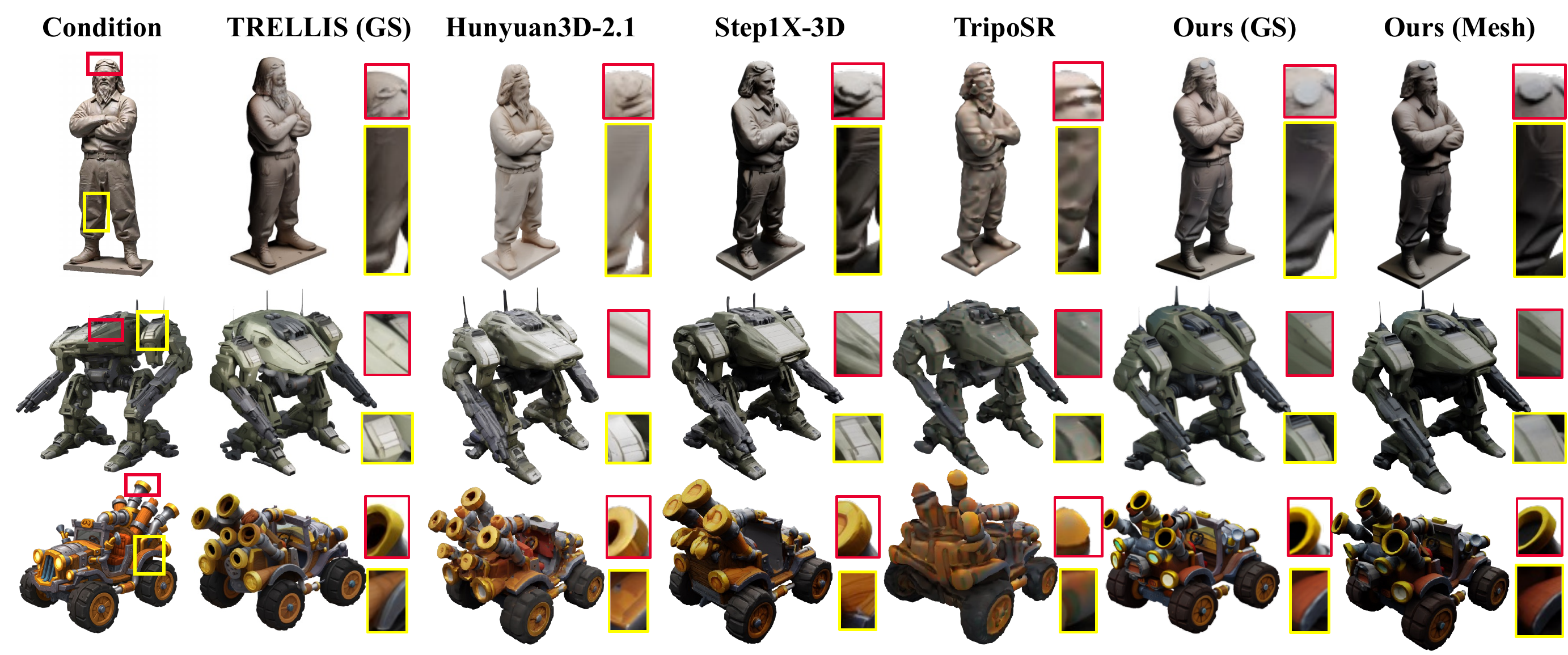}
\caption{Additional qualitative comparisons with other methods.}
\label{fig:comparison_opensource2}
\end{figure}

\begin{figure}[t]
\centering
\includegraphics[width=\linewidth]{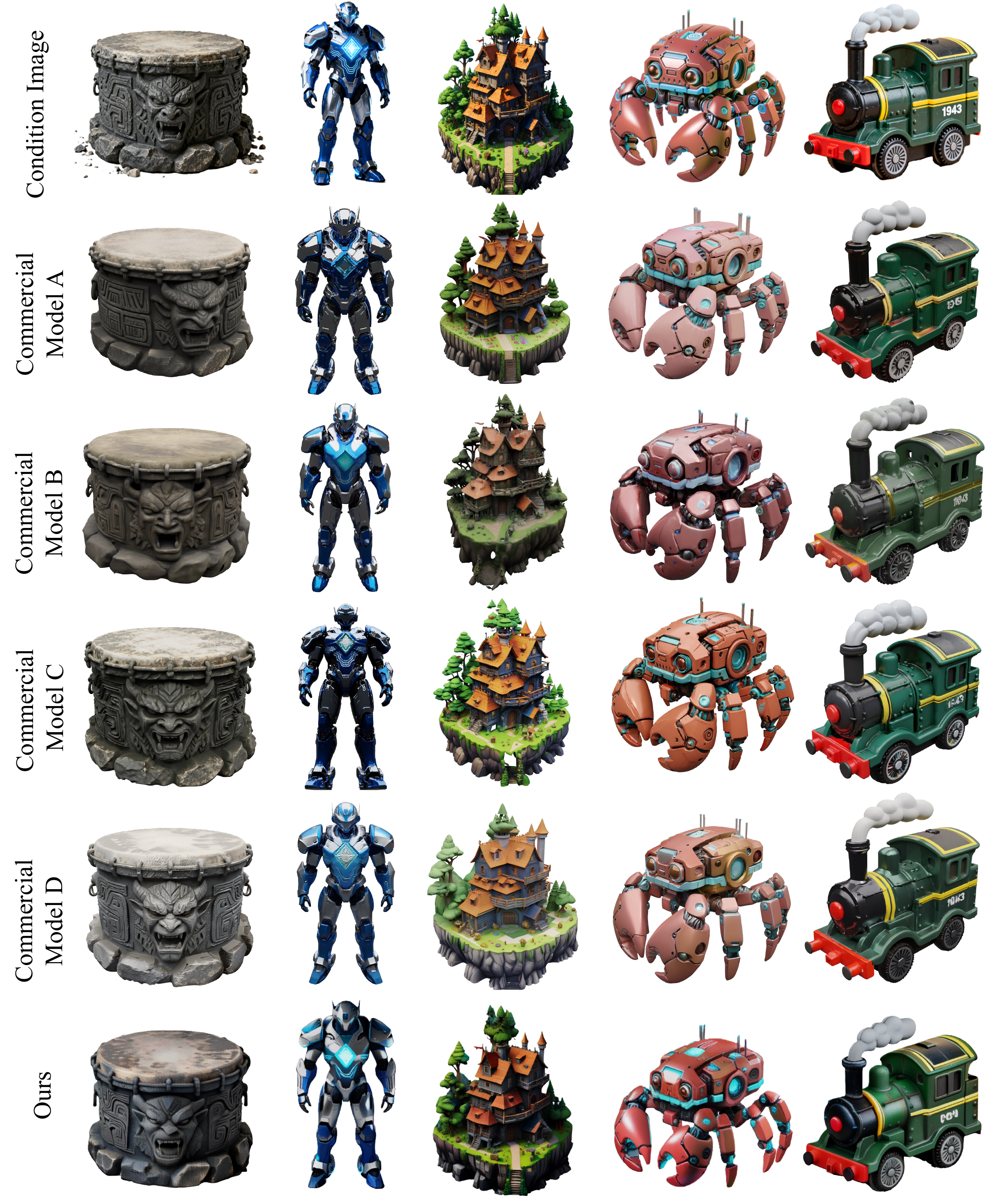}
\caption{Qualitative comparisons with commercial models. Our UniLat3D shows competitive performance even with only publicly available training data.}
\label{fig:comparision_commercial}
\end{figure}

\begin{figure}[t]
\begin{center}
\includegraphics[width=\linewidth]{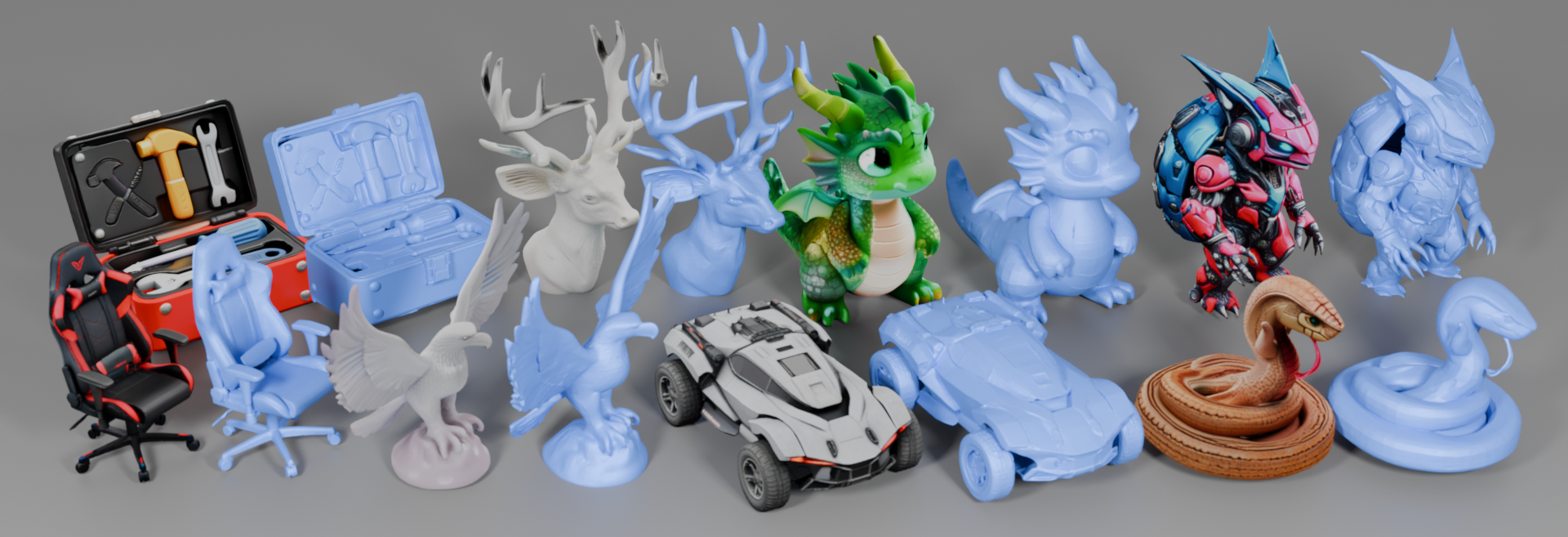}
\end{center}
\vspace{-10pt}
\caption{3D mesh assets generated by our UniLat3D.}
\label{fig:mesh gallery}
\end{figure}

\begin{wrapfigure}{r}{0.3\textwidth}
    \centering
    \vspace{-20pt}
    \includegraphics[width=\linewidth]{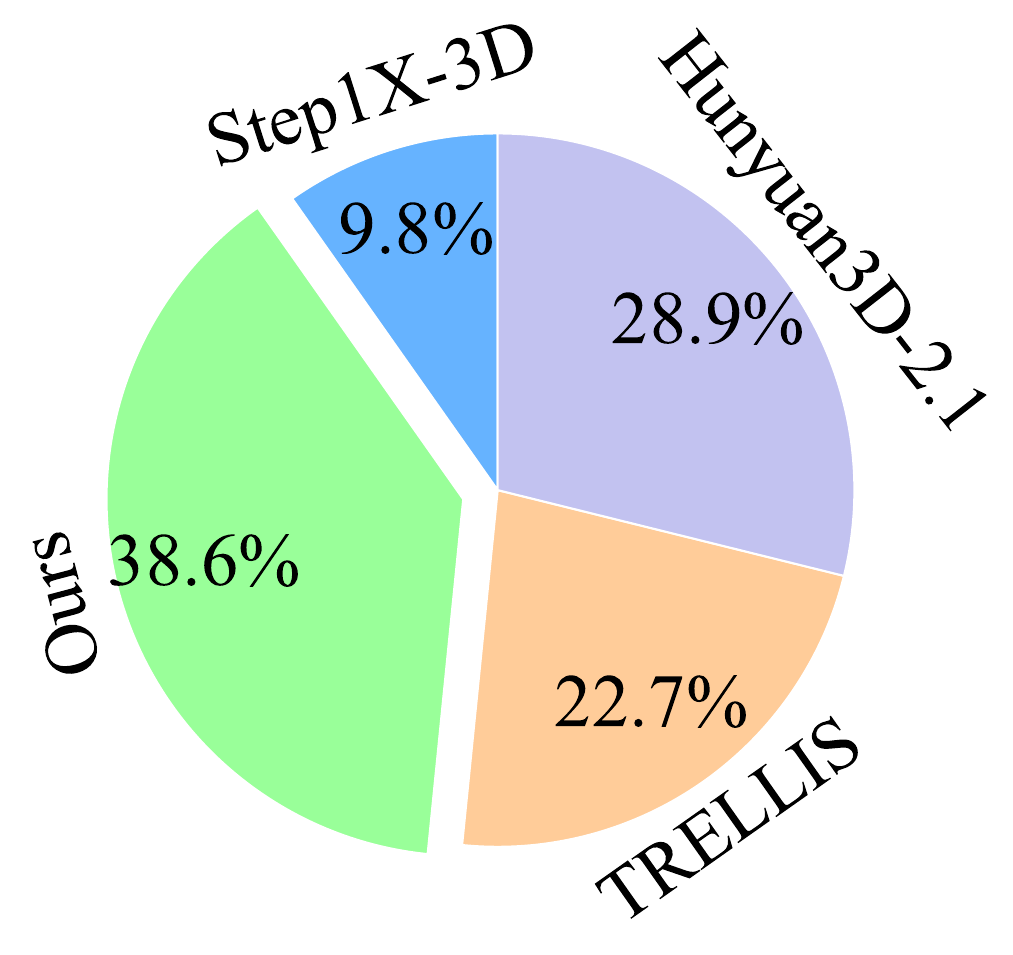}
    \vspace{-25pt}
    \caption{User study on different models.
    }
\label{fig:user study}
\end{wrapfigure}
\subsection{Results}
We provide qualitative comparisons in Fig.~\ref{fig:comparison_opensource} and Fig.~\ref{fig:comparison_opensource2}, where our method achieves competitive generation quality and demonstrates stronger alignment with the conditional image, benefiting from the unified representation. Note that Hunyuan3D-2.1~\citep{hunyuan3d2025hunyuan3d2.1}, Step1X-3D~\citep{li2025step1x}, and TripoSR~\citep{tochilkin2024triposr} only provide mesh-based results. Importantly, Ours, TripoSR, TRELLIS, and Direct3D-S2 are trained exclusively on publicly available datasets, while other methods leverage additional private data. 
% Comparison with other commercial models is also provided by Fig.~\ref{fig:comparision_commercial}.
We also provide qualitative comparisons among some commercial models in Fig.~\ref{fig:comparision_commercial}. Results show that even compared with commercial models, UniLat3D still delivers competitive performance with notably better consistency between the generated 3D content and the input image. 
Fig.~\ref{fig:mesh gallery} displays diverse 3D mesh assets generated by UniLat3D, demonstrating its superior performance in producing high-quality geometry and realistic appearance.

Quantitative evaluations on Toys4k~\citep{toys4k} are reported in Table~\ref{tab:toys4k-appearance}. Additional results on our self-collected complex set are provided in the Table~\ref {tb2:new1k evaluation results}. Compared with other two-stage methods, UniLat3D achieves leading appearance performance, reaching 47.68 in $\text{FD}_\text{DINOv2}$. The CLIP score of 90.87 further demonstrates the effectiveness of UniLat3D in aligning images and 3D assets. In terms of geometry synthesis, our mesh version also achieves competitive results in ULIP~\citep{xue2023ulip}, with a score of 42.69. Beyond accuracy, UniLat3D also demonstrates notable efficiency: 3D Gaussian generation is completed within 8 seconds on a single A100 GPU and can be further reduced to 3 seconds with FlashAttention-3~\citep{shah2024flashattention}. Mesh generation requires 36 seconds, primarily due to the higher resolution with more vertices and longer post-processing compared to TRELLIS, but remains competitive considering the improved output quality. 
% The generated mesh are also provided in Fig.~\ref{fig:mesh gallery}.

Besides, we conducted a user study with 19 participants over 3D assets generated from 23 image prompts. Four models with both geometry and appearance generation are involved. For each prompt, participants judged generated assets by both image alignment and object quality, and chose the overall best case. As shown in Fig.~\ref{fig:user study}, UniLat3D received over 35\% of the votes, outperforming Huanyuan3D-2.1 and other models.

\begin{table}[t]
  \centering
  \caption{Comparisons on the Toys4K dataset. ``\#Params'' denotes the number of model parameters. The ``ULIP'' and ``Uni3D'' metrics are multiplied by $\text{10}^{2}$.}
  \setlength{\tabcolsep}{2mm}
  \label{tab:toys4k-appearance}
  \resizebox{0.95\textwidth}{!}{
  \begin{tabular}{lccc|cc|cc}
    \toprule
    \textbf{Model} & \textbf{Rep.} & \textbf{\#Param.} & \textbf{Time} & \textbf{CLIP$\uparrow$} & \textbf{FD$_\text{DINOv2}$$\downarrow$} & \textbf{ULIP$\uparrow$} & \textbf{Uni3D$\uparrow$} \\
    \midrule
    TripoSR & Mesh & 0.4B  & 13s  & 88.76 & 279.06 & 35.30 & 30.98 \\
    TRELLIS & Mesh & 1.31B & 21s   & 87.81 &  79.52 & 42.51 & 37.67 \\
    TRELLIS & 3DGS & 1.31B & 5s   & 90.70 &  52.54 &  --   &  --   \\
    Stable3DGen$^{\dagger}$$^{\star}$ & Mesh & 2.63B & 4s   &  --   &  --   & 40.33 & 35.98 \\
    Step1X\mbox{-}3D$^{\dagger}$ & Mesh & 4.8B  & 152s & 85.85 & 146.08 & 41.37 & 36.51 \\
    Direct3D\mbox{-}S2$^{\star}$ & Mesh & 2.1B & 185s &  --   &  --   & 41.51 & 36.64 \\
    Hunyuan3D\mbox{-}2.1$^{\dagger}$ & Mesh & 5.3B & 90s  & 88.44 &  74.16 & 42.67 & \textbf{37.74} \\
    \midrule
    Ours & Mesh & 1.58B & 36s & 87.93 & 71.81 & \textbf{42.69} & 37.62 \\
    Ours & 3DGS & 1.55B & 8s  & \textbf{90.87} & \textbf{47.68} & -- & -- \\
    \bottomrule
  \end{tabular}
  }
  \vspace{4pt}
  \begin{minipage}{0.95\textwidth}
    \footnotesize
    $^{\dagger}$ Using proprietary or non\mbox{-}public training data. \\
    $^{\star}$ Only generating geometry without appearance.
  \end{minipage}
  \vspace{-8pt}
\end{table}

\begin{table}[t]
  \centering
    \caption{Comparisons on the self-collected complex test dataset. ``Rep.'' denotes the output representation type, and ``\#Params'' denotes the number of model parameters. The ``ULIP'' and ``Uni3D'' metrics are multiplied by $\text{10}^{2}$.}
    \label{tb2:new1k evaluation results}
  \setlength{\tabcolsep}{2mm}
  \resizebox{0.95\textwidth}{!}{
  \begin{tabular}{lccc|cc|cc}
    \toprule
    \textbf{Model} & \textbf{Rep.} & \textbf{\#Param.} & \textbf{Time} & \textbf{CLIP$\uparrow$} & \textbf{FD$_\text{DINOv2}$$\downarrow$} & \textbf{ULIP$\uparrow$} & \textbf{Uni3D$\uparrow$} \\
    \midrule
    TripoSR & Mesh & 0.4B  & 13s  &  88.00 & 369.86 & 33.61 & 30.44\\
    TRELLIS & Mesh & 1.31B & 21s  &  86.40 &  164.57 & 41.52& 37.30\\ 
    TRELLIS & 3DGS & 1.31B & 5s   & 89.67 &  108.27 & - & - \\
    Stable3DGen$^{\dagger}$$^{\star}$ & Mesh & 2.63B & 4s   &-&-&39.79&35.93\\
    Step1X\mbox{-}3D$^{\dagger}$ & Mesh & 4.8B  & 152s & 84.74 & 210.49 & 40.53 & 36.46\\
    Direct3D\mbox{-}S2$^{\star}$ & Mesh & 2.1B & 185s &-&-&40.77&36.47\\
    Hunyuan3D\mbox{-}2.1$^{\dagger}$ & Mesh & 5.3B & 90s  & 87.41  & 150.39 & 41.70 & \textbf{37.48}\\
    \midrule
    Ours & Mesh & 1.58B & 36s & 86.44 & 149.62 & \textbf{41.71}  & 37.24 \\
    Ours & 3DGS & 1.55B & 8s  & \textbf{89.83} & \textbf{97.22} & - &  -\\
    \bottomrule
  \end{tabular}
  }
  \vspace{4pt}
  \begin{minipage}{0.95\textwidth}
    \footnotesize
    $^{\dagger}$ Using proprietary or non\mbox{-}public training data. \\
    $^{\star}$ Only generating geometry without appearance.
  \end{minipage}
  \vspace{-6pt}
\end{table}

\subsection{Ablation Study}
\begin{wraptable}{r}{0.6\linewidth}
  \centering
  \vspace{-50pt}
  \caption{VAE reconstruction results with latents of different resolutions.}
  \small
  \begin{tabular}{lcccc}
    \toprule
    \textbf{Model} & \textbf{Res.}  & \textbf{PSNR$\uparrow$} & \textbf{SSIM$\uparrow$} & \textbf{LPIPS$\downarrow$}  \\
    \midrule
    TRELLIS (Mesh) & $64^3$ & 31.91 &  97.44 & 0.0328 \\
    Ours (Mesh) & $16^3$ & 32.35 & 98.03  & 0.0305 \\
    \midrule
    TRELLIS (GS) & $64^3$ & 34.74 & 98.52  & 0.0146 \\
    Ours (GS) & $8^3$ & 33.51 & 98.13  & 0.0200 \\
    Ours (GS) & $16^3$ & 34.80 & 98.49  & 0.0158 \\
    Ours (GS) & $32^3$ & \textbf{34.92} & \textbf{98.53}  & \textbf{0.0145} \\
    \bottomrule
  \end{tabular}
  \label{tb:ablation_vae}
\end{wraptable}

\paragraph{Resolution of Latents.}
We explore the latent space of reconstruction quality in Uni-VAE. We train Uni-VAE at different latent resolutions, including $8^3$, $16^3$, and $32^3$. As shown in Table~\ref{tb:ablation_vae}, higher UniLat resolutions lead to better reconstruction results. Note that our Uni-VAE achieves similar or even better reconstruction performance than TRELLIS with smaller resolutions. In our experiments, when training the flow Transformer at a higher resolution of 32, the computational cost increases evidently. We would explore more efficient approaches on flow Transformers for higher resolutions in future works, \eg, block-wise computation and lightweight attention. 

\begin{table}[h!]
  \centering
  \caption{Ablation study on the visual encoder for condition images.}
\setlength{\tabcolsep}{1mm}
  \small
  \begin{tabular}{lccc}
    \toprule
    \textbf{Model} & \textbf{Cond. Encoder}  & \textbf{CLIP$\uparrow$} & \textbf{$\text{FD}_{\text{dinov2}}$$\downarrow$} \\
    \midrule
    Ours & DINOV2 & \textbf{90.83}   & 52.58 \\
    Ours & DINOV3 & 90.60 & \textbf{49.90} \\
    \bottomrule
  \end{tabular}
  \label{tab:ablation_encoder}
\end{table}
\paragraph{Visual Encoder of Condition Images}
Recently, DINOv3~\citep{simeoni2025dinov3} emerges as a strong visual encoder model that could extract high-quality details from the image. We compare the performance between DINOv2 and DINOv3 for encoding condition images. Flow models with different visual encoders are trained for 500 iterations and tested on Toys4K. In our experiments, the flow Transformer with the DINOv3 encoder shows better quality on complex object generation, which leads to a better $\text{FD}_{\text{dinov2}}$ result as shown in Table~\ref{tab:ablation_encoder}.

\section{Discussion \& Conclusion}
We propose a novel 3D generation framework -- UniLat3D to achieve high-quality 3D asset generation in seconds with a single-stage flow model. Apart from that the proposed method unifies geometry and appearance in a single, concise framework, it achieves quite competitive performance compared with popular two-stage methods. We expect our exploration to provide a more convenient and extensible choice to the 3D generation field, \eg, further unifying object and scene generation with the compact unified representation, extending UniLat to 4D representations, and integrating UniLat into large multimodal models \etc.

However, the UniLat3D model implemented in this paper is still a preliminary exploration. The training data we used just follows TRELLIS, totally from public datasets. Injecting more high-quality data for training will undoubtedly improve the performance and may further scale up the model. Exploring more efficient designs on the flow model would adapt to higher resolutions of latents, leading to more detailed generation results.

\section*{Acknowledgement}
We would like to thank Junjie Wang and Zhikuan Bao for their valuable contributions to this project. We are also grateful to Jinfeng Yao and Lianghui Zhu for their valuable input during the initial stages of the project.

\appendix

\bibliography{2026_conference}
\bibliographystyle{2026_conference}
\clearpage

\section{Appendix}
\subsection{Model Architecture}
\label{app_subsec:model_architecture}
In this section, we mainly provide the model architecture about our Uni-VAE $\{\mathcal{E}_{uni}, \mathcal{D}_{uni}\}$ and UniLat generation model $\mathcal{F}$.

\subsubsection{Uni-VAE}
For the sparse encoder $\mathcal{M}_{sparse}$, we mainly follow TRELLIS's configurations to build a sparse Transformer. For the dense encoder $\mathcal{M}_{dense}$, a set of conv3D layers is used as the main architecture. The settings of $\mathcal{E}_{sparse}$, $\mathcal{D}_{up}$ are shown in Table~\ref{tab:hyperparam:dense} and details of $\mathcal{E}_{uni}$ are provided in Table~\ref{tab:structure_vae}.

\begin{table}[ht]
  \centering
  \caption{Model details of Uni-VAE modules $\mathcal{M}_{dense}, \mathcal{D}_{up}$. ``Channels'' denotes model channels after each up/downsampled convolution layer.}
  \vspace{0.5em}
  \label{tab:hyperparam:dense}
  \begin{tabular}{lcc}
    \toprule
    \textbf{Model} & \textbf{ResBlocks} & \textbf{Channels} \\
    \midrule
    $\mathcal{E}_{sparse}$ & 4 & $[32,128,512]$ \\
    $\mathcal{D}_{up}$     & 4 & $[512,128,32]$ \\
    \bottomrule
  \end{tabular}
\end{table}

\begin{table}[ht]
  \centering
  \caption{Model details of Uni-VAE modules $\mathcal{M}_{sparse}, \mathcal{D}_{gs,mesh}$.}
    \resizebox{\textwidth}{!}{
  \begin{tabular}{lcccccccc}
    \toprule
    \textbf{Model}   & \textbf{Latent Res.}   & \textbf{Model Channels}&\textbf{Latent. Channels} & \textbf{Blocks} & \textbf{Attn. Heads} & \textbf{Window Size}\\
    \midrule
    $\mathcal{M}_{sparse}$, $\mathcal{D}_{sparse}$  & 64 & 768 & 8 & 12&12&8 \\
    \bottomrule
  \end{tabular}
  }
  \label{tab:structure_vae}
\end{table}

\subsubsection{UniLat Flow Transformer}
Structure details about our UniLat flow Transformer $\mathcal{F}_{uni}$ are provided in the Table~\ref{Table:Structure_UniFlow}. The main architecture of $\mathcal{F}_{uni}$ is similar to TRELLIS's sparse structure flow Transformer. The input noise $\epsilon$ would be flattened to 1D tensors. Positional encoding is applied to a flattened tensor, and it would be fed to Transformer blocks with self\&cross-attention layer and modulated by condition signal \& timestamps. Finally, the flattened tensor would be unpatchified to 3D results, the shape is the same as $\epsilon$.
\begin{table}[hb]
  \centering
  \caption{Model details of UniLat3D flow Transformer.}
    \resizebox{\textwidth}{!}{
  \begin{tabular}{lccccccccc}
    \toprule
    \textbf{Model} & \textbf{Params}  & \textbf{Latent Res.}  & \textbf{Latent Channels} & \textbf{Model Channels}&\textbf{Cond. Channels} & \textbf{Blocks} & \textbf{Attn. Heads}\\
    \midrule
    $\mathcal{F}_{uni}$ & 1.30B  & 16 & 32 & 1280 & 1280 & 36 & 32 \\
    \bottomrule
  \end{tabular}
  \label{Table:Structure_UniFlow}
  }
\end{table}
\end{document}

%% file: math_commands.tex
\usepackage{amsmath,amsfonts,bm}

\def\eqref#1{equation~\ref{#1}}
\def\1{\bm{1}}

\DeclareMathAlphabet{\mathsfit}{\encodingdefault}{\sfdefault}{m}{sl}
\SetMathAlphabet{\mathsfit}{bold}{\encodingdefault}{\sfdefault}{bx}{n}